\documentclass[12pt]{article}
\usepackage{amsmath}
\usepackage{amssymb}
\usepackage{graphics}

\begin{document}
\title{On the predictability of Rainfall in Kerala\\
       An application of ABF Neural Network}
\author{Ninan Sajeeth Philip
\\
and\\
K. Babu Joseph
\\
Cochin University of Science and Technology, \\
Kochi-682022, Kerala State, India.
}
\date{Draft for ISDA2001}
\maketitle
\begin{abstract}
Rainfall in Kerala State, the southern part of Indian Peninsula in particular is
caused by the two 
monsoons and the two cyclones every year. In general, climate and rainfall are highly 
nonlinear phenomena in nature giving rise to what is known as the `butterfly effect'. We 
however attempt to train an ABF neural network on the time series rainfall data and show for the
first time that
in spite of the fluctuations resulting from the nonlinearity in the system, the
trends in the rainfall pattern in this corner of the globe have remained unaffected over the 
past 87 years from 1893 to 1980. We also successfully filter out the chaotic part of the
system and illustrate that its effects are marginal over long term predictions.
\end{abstract}
\section{Introduction}
Although technology has taken us a long way towards better living standards, we still have a
significant dependence on nature. Rain is one of nature's greatest gifts and in third world 
countries
like India, the entire agriculture depends upon rain. It is thus a major concern
to identify any trends for rainfall to deviate from its periodicity, which would 
disrupt the economy of 
the country. This fear has been aggravated due to the threat by the global warming and
green house effect. The present study has a soothing effect since it concludes that in spite of 
short term  fluctuations, the general pattern of rainfall in Kerala has
not undergone major deviations from its pattern in the past.

The geographical configuration of India with the three oceans, namely Indian Ocean, Bay of
Bengal and the Arabian sea bordering the peninsula gives her a climate system with
two monsoon seasons and two cyclones inter-spersed with hot and cold weather seasons. The 
parameters that are required to predict the rainfall are enormously complex and subtle so that
the uncertainty in a prediction using all these parameters is enormous even for a short
period. The period over which a prediction may be made is generally termed the event horizon
and in best results, this is not more than a week's time. Thus it is generally said that the
fluttering wings of a butterfly at one corner of the globe may cause it to produce a tornado
at another place geographically far away. This phenomenon is known as the 
\textit{butterfly effect.}

The objective of this study is to find out how well the periodicity in these patterns may be
understood using a neural network so that long term predictions can be made. This would
help one to anticipate with some degree of confidence the general pattern of rainfall 
for the coming years. To evaluate the performance of the network, we train it on
the rainfall data corresponding to a certain period in the past and cross validate the 
prediction made by the network over some other period. A difference diagram\footnote{This is the
diagram showing the deviation of the predicted sequence from the actual rainfall pattern.} 
is plotted to estimate the extent of deviation between the 
predicted and actual rainfall. However, 
in some cases, the cyclone may be either 
delayed or rushed along due to some hidden perturbations on the system, for example
an increase in solar activity. (See figure \ref{figure2}.)  These effects would appear as spikes
in the difference diagram. The information from the difference diagram is insufficient 
to identify the exact source of such spike formation. It might have resulted from slight
perturbations from unknown sources or could be due to an inaccurate modeling of the system using
the neural network.  We thus use a standard procedure in statistical 
mechanics that can quantitatively estimate the fluctuations\cite{ESRGopal}. This is to estimate
the difference in the Fourier Power Spectra (FPS) of the predicted and actual sequences. 
In the FPS, the power corresponding
to each frequency interval, referred to as the \textit{spectral density} gives a quantitative
estimate of the deviations of the model from reality. Rapid variations would contribute to
high frequency terms and slowly varying quantities would correspond to low frequency terms in
the power spectra. 

The degree of information that may be extracted from the FPS is of great significance. If the
model agrees with reality, the difference of the power spectra 
( {\em hereafter referred to as the residual FPS} ) should enclose minimum power in the
entire frequency range. An exact model would produce no difference and thus no residual FPS
pattern. If there are some prominent frequency components in the residue, that could indicate
two possibilities; either the network has failed to comprehend the periodicity, or that there
is a new trend in the real world which did not exist in the past. One can test whether the same
pattern exists in the residual FPS produced on the training set and confirm whether it is a new
trend or is the drawback of the model. 

A random fluctuation would be indicated in the residual FPS by amplitudes at all frequency 
values as in the case of `white' noise spectra.
Here again, how much power is enclosed within the FPS gives a quantitative estimate of the 
perturbations. A low amplitude fluctuation can happen due to so many reasons. But its
effect on the overall predictability of the system would be minimal. If however, the 
residual FPS
encloses a substantial power from the actual rainfall spectrum, the fluctuations could be 
catastrophic.

In this study, the results indicate that the perturbations produced by the environment on the
rainfall pattern of Kerala state in India is minimal and that there is no evidence to envisage a
significant deviation from the rainfall patterns prevailing here.

The general outline of the paper is as follows. In section \ref{ABFNN} we
present a brief  outline of the Adaptive Basis Function Neural Network
(ABFNN), a variant of the popular back-propagation algorithm. In 
section \ref{Setups}, the experimental set up is explained followed by
the results and  concluding remarks in section \ref{Results}.

\section{Adaptive Basis Function Neural Networks}\label{ABFNN}

It was shown in \cite{NSPKBJ} that a variant of the back-propagation algorithm (backprop)
known as the
Adaptive Basis Function Neural Network performs better than the standard backprop networks
in complex problems. The ABFNN works on the principle that the neural network always attempts
to map the target space in terms of its basis functions or node functions. In standard 
backprop networks, this function is a fixed sigmoid function that can map between zero and
plus one or between minus one and plus one the input applied to it from minus infinity to plus
infinity. It has many attractive properties that make the backprop an efficient tool in a wide
variety of applications. However serious studies conducted on the backprop algorithm have shown
that in spite of its widespread acceptance, it systematically outperforms other classification
procedures \textit{only} when the targeted space has a sigmoidal shape \cite{Kraaijveld}. This 
implies that one should \textit{choose} a basis function such that the network may represent 
the target space as a nested sum of products of the input parameters in terms of the basis 
function. The ABFNN thus starts with the standard sigmoid basis function and alters its 
nonlinearity by an algorithm similar to the weight update algorithm used in backprop.

Instead of the standard sigmoid function, ABFNN opts for a variable sigmoid function defined as
\begin{equation}
O_f = \frac{a+tanh(x)}{1+a}
\end{equation}
Here \(a\) is a control parameter that is initially set to unity and is modified along with the
connection weights along the negative gradient of the error function. It is claimed in 
\cite{NSPKBJ} that such a modification could improve the speed and accuracy with which the 
network could approximate the target space.

The error function is computed as:
\begin{equation} 
E=\sum _{k}\frac{\left( O_{k}-O_{k}^{*}\right) ^{2}}{2} 
\end{equation}
{\noindent{with $O_{k}$ representing the network output and \( O_{k}^{*} \)
representing the target output value.}}

With the introduction of the control parameter, the learning algorithm may be summarized by the
following update rules. It is assumed that each node, \(k\), has an independent node function 
represented by $ a_{k}$. For the output layer nodes, the updating is done
by means of the equation:
\begin{equation} 
\qquad \Delta a_{k}=-\beta \left( O_{k}-O_{k}^{*}\right) \frac{1-O_{k}}{1+a_{k}} 
\end{equation}
where $\beta$ is a constant which is identical to the learning parameter in the weight update
procedure used by backprop.

For the first hidden layer nodes, the updating is done in accordance with the equation:
\begin{equation}
 \Delta {\textit {a}}_{k-1}=-\sum _{i}w_{ij}\beta \left( O_{k}-O_{k}^{*}\right) \left( 1-O_{k}\right) \left[ \left( 1+{\textit {a}}_{k}\right) +O_{k}\left( 1+{\textit {a}}_{k}\right) \right] \frac{\partial O_{k}}{\partial O_{k-1}} 
 \end{equation}
Here \(w_{ij} \) is the connection weight for the propagation of the output from node \(i\) to
node \(j\) in the subsequent layer in the network.

The introduction of the control parameter results in a slight modification 
to the weight update
rule ($ \Delta w_{ij} $) in the computation of the second partial derivative term
 \( \frac{\partial O_{j}}{\partial I_{j}} \)
in:
\begin{equation}
 \Delta w_{ij}=-\beta \frac{dE}{dw_{ij}} =-\beta \frac{\partial E}{\partial O_{j}}\frac{\partial O_{j}}{\partial I_{j}}\frac{\partial I_{j}}{\partial w_{ij}} 
 \end{equation}
as:
 \begin{equation} 
 \frac{\partial O_{j}}{\partial I_{j}}=\left[ \left( 1-a_{j}\right) +\left( 1+a_{j}\right) O_{j}\right] \left( 1-O_{j}\right)
  \end{equation}
The algorithm does not impose any limiting values for the parameter \(a\) and it is assumed that 
care is taken to avoid division by zero.  

\section{Experimental setup}\label{Setups}

In pace with the global interest in climatology, there has been a rapid updating of resources in
India also to access and process climatological database. There are various data acquisition 
centers in the country that record daily rainfall along with other measures such as sea surface
pressure, temperature etc. that are of interest to climatological processing. These centers are
also associated to the World Meteorological Organization (WMO). The database used for this study
was provided by the Department of Atmospheric Sciences of Cochin University of Science and
Technology, a leading partner in the nation wide climatological study centers.

The database consists of the rainfall data from Trivandrum in Kerala, situated at 
latitude-longitude pairs ($8^o 29^\prime$ N - $76^o57^\prime$ E). Although the rainfall data 
from 1842 were recorded,
there were many missing values in between that we had to restrict to periods for which a
continuous time series was available. This was obtained for the period from 1893 to 1980.

For training our network, the Trivandrum database from 1893 to 1933 was 
used. Since rainfall has an yearly periodicity, we started with a network having 12 input 
nodes. It was observed that the network accuracy would systematically improve as we increased the
number of input nodes from 12 to 48 covering the data corresponding to 4 years. Any increase
in input nodes resulted in poorer representations. Further experimentation showed that it was
not necessary to include information corresponding to the whole year, but a 3 month 
information centered over the predicted month of the fifth year in each of the 4 previous years
would give good generalization properties. We thus finalized our network with 12 input nodes
each for the 3 months input data over 4 years, 7 hidden nodes and one output node. Thus, based
on the information from the four previous years, the network would predict the amount of rain
to be expected in each month of the fifth year.

 The training was carried out until the root mean-square error stabilized
to around 0.085 over the training data(40 years from 1893-1933). The ABFNN converged to this value in around 1000 
iterations starting from random weight values and took less than 10 minutes on a 300MHz PC
using Celeron processor and Linux operating system. 

After training, the data corresponding to the entire 87 years of rainfall in Trivandrum city was 
presented to the network. The output from the network was compared with the actual data in the 
time series.  In addition to visual comparison, the spectral analysis was also done on the test dataset
to obtain a quantitative appreciation of the performance of the network.

\section{Results and Discussion}\label{Results}

\begin{figure}[ht]

{\centering \resizebox*{0.95\columnwidth}{0.5\textheight}{\includegraphics{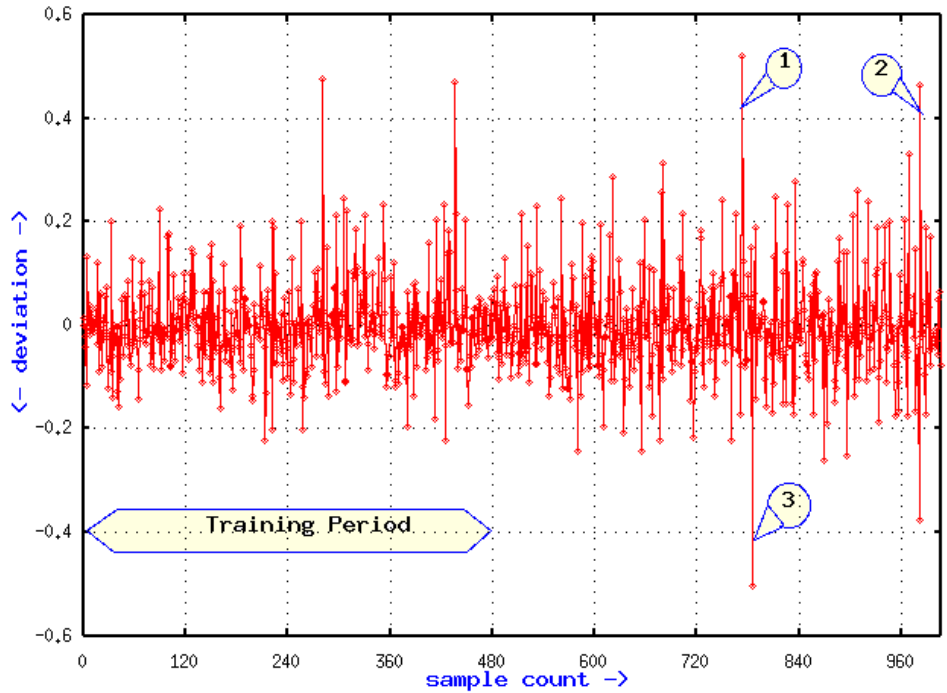}}}

\caption{\label{figure1} The deviation of the predicted sequence from the actual time series
data is shown. It may be noted that the deviations in the training period and the rest of the
series are comparable in magnitude. The larger spikes are mainly due to the delay in the actual
commencement of the monsoon or cyclone. Label 1 indicates an example where the monsoon extended
into the following month giving more rainfall than anticipated. Label 2 in the figure 
represent an example where the monsoon was actually delayed by one month than was anticipated
thus producing two spikes, one negative and the other positive.  Label 3 corresponds to a 
negative spike caused by the change in trend from the predicted increase in rain in the following
year.}
\end{figure}

\begin{figure}[ht] 

{\centering \resizebox*{0.95\columnwidth}{0.5\textheight}{\includegraphics{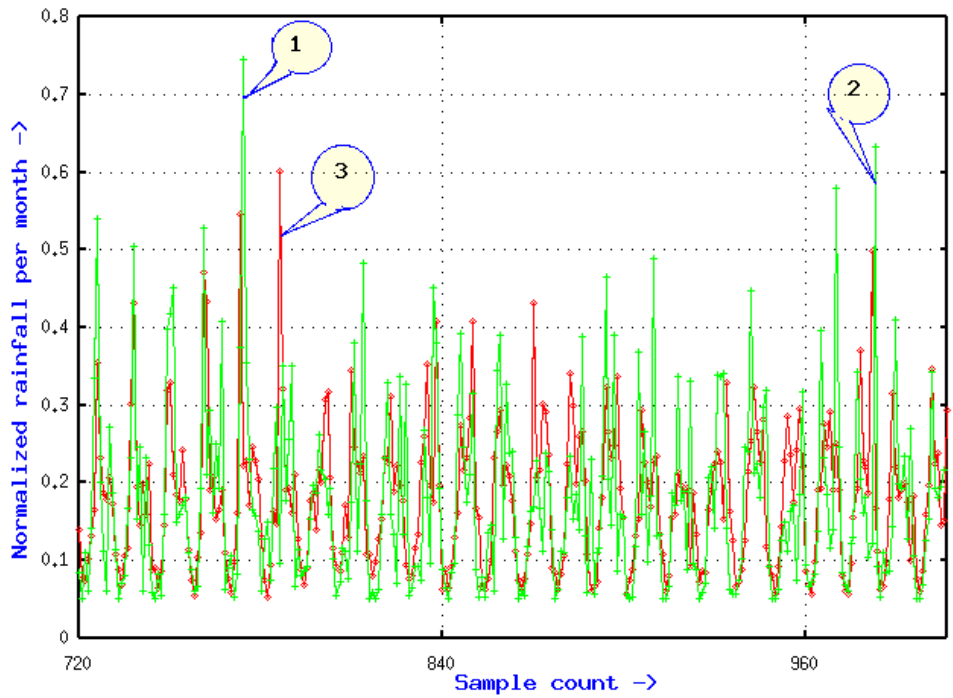}}}

\caption{\label{figure2} An overlap of the predicted and the actual time series data in the
region corresponding to the spikes in figure 1 is shown. The labels indicate the reason
for the spikes seen in figure 1.}
\end{figure}

In figure \ref{figure1} we show the difference pattern produced by the network over the entire
87 year period on the Trivandrum database. The training period is also shown. It may be noted
that the deviations of the predicted signal from the actual in the entire dataset falls 
within the same magnitude range. The root mean square error on the independent test dataset was 
found to be around 0.09. The two positive spikes visible in
the plot corresponding to the two instances where the rainfall was delayed with reference to the 
month for which it was predicted are shown in figure \ref{figure2}. 
 
The cause for the 
negative spike in figure \ref{figure1} is also seen in figure \ref{figure2} as due to the
deviation of the time series from the predicted upward trend. Factors such as the 
El-Nino southern
oscillations (ENSO) resulting from the pressure oscillations between the tropical Indian 
Ocean and the tropical Pacific Ocean and their quasi periodic oscillations are observed to
be the major cause of these deviations in the rainfall patterns\cite{Chowdhury91}.

\begin{figure}[ht] 

{\centering \resizebox*{0.95\columnwidth}{0.5\textheight}{\includegraphics{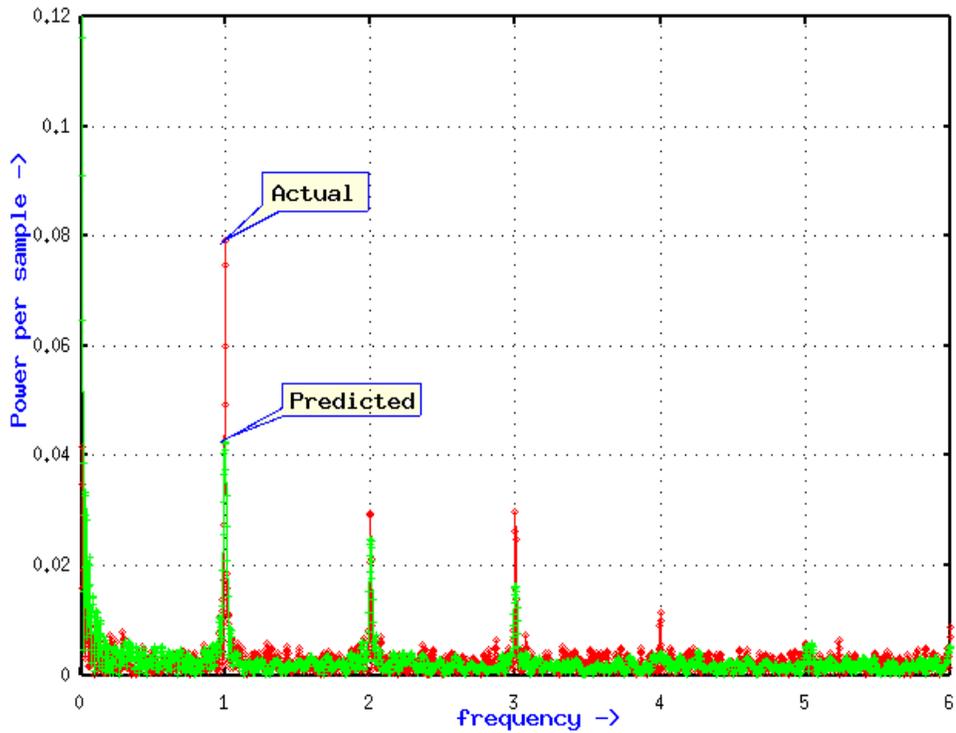}}}

\caption{\label{figure3} The Fourier power spectra corresponding to the actual and the predicted
rainfall for the entire 87 years. The sampling frequency does not have
much significance here since the sample values represent the total rain
received in each month. It was taken to be twelve to  associate itself
with the 12 months of the year. Since FPS is symmetric over half the
sampling  frequency, we show here only values from 0 to 6.}
\end{figure}

\begin{figure}[ht] 

{\centering \resizebox*{0.9\columnwidth}{0.4\textheight}{\includegraphics{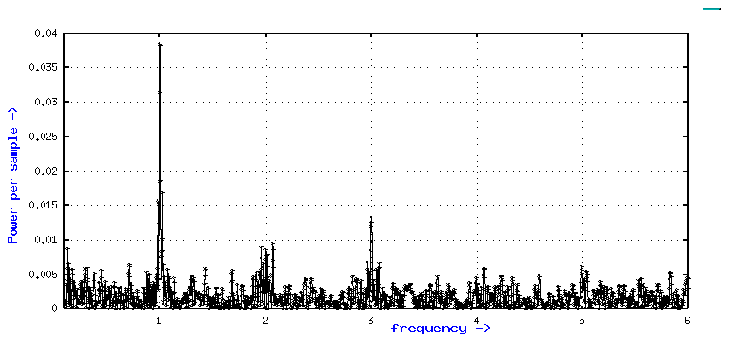}}}

{\centering \resizebox*{0.9\columnwidth}{0.4\textheight}{\includegraphics{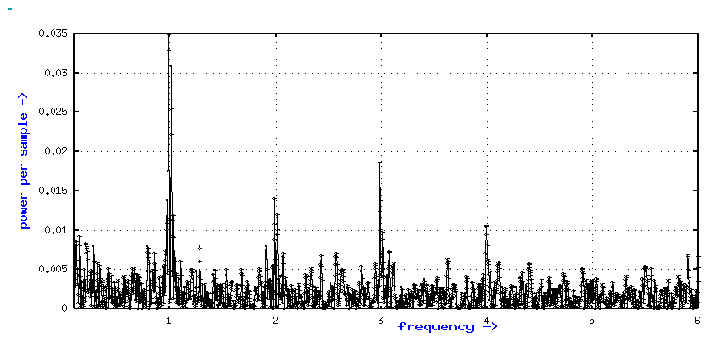}}}

\caption{\label{figure4} The residual FPS corresponding to the 
predicted and actual rainfall in the years following the training period is
shown on top. The spectral difference from a Fourier analysis
perspective (the difference of the FPS in the training period and the
period corresponding to the top figure) is presented in the bottom figure.
Note that the spectral differences are minimum in the top figure. This is
to say that the deviations observed in the actual spectra in the bottom
figure followed from the trend associated with the training period. }
\end{figure}

To obtain a quantitative appreciation of the learning algorithm, we resort
to the spectroscopic analysis of the predicted and actual time series.
The  average power per sample distributed in the two sequence is
shown in figure \ref{figure3}. It is seen that the two spectra compare
very well. To identify the differences, the deviation  of the FPS
(residual FPS) of the predicted  sequence from the actual in the months
following the training period was plotted \footnote{This data corresponds to the independent 
test period.}. For comparison, The deviation
of the FPS of the actual rainfall patterns in this period from that of the 
training period was also plotted. The resulting graphs are shown in
figure \ref{figure4}. In attempting to predict the evolution of the system
in the months following the training period, it is to be noted that the
deviation of the spectra  of the predicted sequence for this period is
less than that of the Fourier spectra corresponding to the training
period.  This is because  the network attempts to \textit{learn} the
behavior of a system where as Fourier analysis means only a
transformation from time domain to frequency domain. Also Fourier
analysis assumes the system to be stationary, which is  not true in the
case of rainfall phenomena. Comparing the FPS of the network output
corresponding to the training period showed the existence of residual
frequency components at 1 Hz and 3 Hz, as seen in the residual FPS  of
figure \ref{figure4}. They represent the error bars or the inherent
uncertainty to be expected in the prediction of the quantum of rainfall
due to those oscillations. The information for \textit{learning} such
fluctuations is not in the time series data. Future studies shall include
other climatic parameters such as temperature, sea surface pressure,
solar activity etc. in conjunction with the time series information in an
attempt to reduce the error bars in the prediction. The rest of the
spectra is spread over the entire   frequency range producing random
fluctuations in the rainfall phenomena.

Thus it is to be concluded that other than minor fluctuations, the power factor in the residual 
FPS is insignificant 
to expect any major deviations in the rainfall pattern prevailing in the country.

It is also to be noted that the ABF neural network is a better tool than the popular 
Fourier analysis methods for predicting long term rainfall behavior. The learning ability of
the network can give a concise picture of the actual system where as the Fourier spectra is 
only a transformation from time domain to frequency domain, assuming that the time domain sequence
is stationary. It thus fails to represent the dynamics of the system which is inherent in most
natural phenomena and rainfall in particular.

\textbf{Acknowledgment}

The authors wish to express their sincere thanks to Professor K. Mohankumar of the Department of
Atmospheric Sciences of the Cochin University of Science and Technology for providing us with the
rainfall database and for useful discussions. The first author would like to thank Dr. Moncy V
John of the Department of Physics, Kozhencherri St. Thomas college for long hours of discussions.

\end{document}